\documentclass{llncs}
\usepackage{graphicx} 
\usepackage{subfigure}
\usepackage{amsmath}
\usepackage{amssymb}
\usepackage{epstopdf}
\usepackage{algorithm}
\usepackage{algorithmic}
\usepackage{float}
\usepackage{hyperref}

\usepackage{hyperref}

\newtheorem{mydef}{Theorem}
\newtheorem{mydef2}{Theorem}

\newenvironment{proofsketch}{\par{\noindent \bf Proof Sketch:}}{\qed\par}
\newtheorem{lemmi}{Lemma}

\begin{document}

\title{Representative Selection in Non Metric Datasets}
\author{Elad Liebman\inst{1} \and Benny Chor\inst{2} \and Peter Stone\inst{1}}

\institute{Computer Science Department \\ The University of Texas at Austin \\ 2317 Speedway, Stop D9500 \\ Austin, TX 78712  \and School of Computer Science \\ Tel Aviv University \\ P.O.B. 39040, Ramat Aviv \\ Tel Aviv 69978}

\maketitle

\begin{abstract} 
This paper considers the problem of representative selection: choosing a subset of data points from a dataset that best represents its overall set of elements. This subset needs to inherently reflect the type of information contained in the entire set, while minimizing redundancy. For such purposes, clustering may seem like a natural approach. However, existing clustering methods are not ideally suited for representative selection, especially when  dealing with non-metric data, where only a pairwise similarity measure exists. In this paper we propose $\delta$-medoids, a novel approach that can be viewed as an extension to the $k$-medoids algorithm and is specifically suited for sample representative selection from non-metric data. We empirically validate $\delta$-medoids in two domains, namely music analysis and motion analysis. We also show some theoretical bounds on the performance of $\delta$-medoids and the hardness of representative selection in general.  
\end{abstract}


\section{Introduction}
\label{intro}

Consider the task of a teacher who is charged with introducing his
class to a large corpus of songs (for instance, popular western music
since 1950). In drawing up the syllabus, this teacher will need to
select a relatively small set of songs to discuss with his students
such that 1) every song in the larger corpus is represented by his
selection (in the sense that it is relatively similar to one of the
selected songs) and 2) the set of selected songs is small enough to
cover in a single semester. This task is an instance of the
\emph{representative selection} problem. Similar challenges often
arise in tasks related to data summarization and modeling. For
instance, finding a characteristic subset of Facebook profiles out of
a large set, or a subset of representative news articles from the entire set of news information gathered during a single day from many different sources.

On its surface, representative selection is quite similar to
clustering, a more widely studied problem in unsupervised learning.
Clustering is one of the most widespread tools for studying the
structure of data. It has seen extensive usage in countless research
disciplines. The objective of clustering is to partition a given data
set of samples into subsets so that samples within the same subset are
more similar to one another than samples belonging to different
subsets. Several surveys of
clustering techniques can be found in the
literature~\cite{jain1999data,xu2005survey}.

The idea of reducing a full set to a smaller set of representatives has been suggested before in specific contexts, such as clustering xml documents  \cite{de2003distance} or dataset editing \cite{eick2004using}, and more recently in visual \cite{hadi2006video,chu2008automatic} and text summarization \cite{nenkova2011automatic}. It has also been discussed as a general problem in \cite{wang2013beyond}. These recurring notions can be formalized as follows. Given a large set of examples, we seek a minimal subset that is rich enough to encapsulate the entire set, thus achieving two competing criteria - maintaining a representative set as small as possible, while satisfying the constraint that all samples are within $\delta$ from some representative. In the next subsections we define this problem in more exact terms, and motivate the need for such an approach.

While certainly related, clustering and representative selection are not the same problem. A seemingly good cluster may not necessarily contain a natural single representative, and a seemingly good partitioning might not induce a good set of representatives. For this reason, traditional clustering techniques are not necessarily well suited for representative selection. We expand on this notion in the next sections.

\subsection{Representative Selection: Problem Definition}
\label{probdef}
Let $S$ be a data set, $d:S \times S \rightarrow\mathbb{R}^+$ be a distance measure (not necessarily a metric), and $\delta$ be a distance threshold below which samples are considered sufficiently similar. We are tasked with finding a \emph{representative} subset $C \subset S$ that best encapsulates the data. We impose the following two requirements on any algorithm for finding a representative subset:

\begin{itemize}
\item {\bf \emph{Requirement $1$:}} The algorithm must return a subset $C \subset S$ such that for any sample $x \in S$, there exists a sample $c \in C$  satisfying $d(x, c) \leq \delta$.
\item {\bf \emph{Requirement $2$:}} The algorithm cannot rely on a metric representation of the samples in $S$.
\end{itemize}
To compare the quality of different subsets returned by different algorithms, we measure the quality of encapsulation by two criteria:

\begin{itemize}
\item {\bf \emph{Criterion $1$:}} $|C|$ - we seek the \emph{smallest} possible subset $C$ that satisfies \emph{Requirement $1$:}.

\item {\bf \emph{Criterion $2$:}} We would also like the representative set to best fit the data \emph{on average}. Given representative subsets of equal size, we prefer the one that minimizes the average distance of samples from their respective representatives.
\end{itemize}

{\setlength{\parindent}{0cm}Criteria $1$ and $2$ are applied on a representative set \emph{solution}. In addition, we expect the following desiderata for a representative selection \emph{algorithm}.

\begin{itemize}
\item {\bf \emph{Desideratum $1$:}} We prefer representative selection algorithms which are \emph{stable}. Let $C_1$ and $C_2$ be different representative subsets for dataset $S$ obtained by two different runs of the same algorithm. Stability is defined as the overlap $\frac{|C_1 \cap C_2|}{|C_1 \cup C_2|}$. The higher the expected overlap is, the more stable the algorithm is. This desideratum ensures the representative set is robust to randomization in data ordering or the choices made by the algorithm. 

\item {\bf \emph{Desideratum $2$:}} We would like the algorithm to be efficient and to scale well for large datasets.
\end{itemize}

Though not crucial for correctness, the first desideratum is useful for consistency and repeatability. We further motivate the reason for desideratum 1 in Appendix B, and show it is reasonably attainable.

The representative selection problem is similar to the $\epsilon$-covering number problem in metric spaces \cite{zhang2002covering}. The $\epsilon$-covering number measures how many small spherical balls would be needed to completely cover (with overlap) a given space. The main difference is that in our case we also wish the representative set to closely fit the data (Criterion $2$). Criteria $1$ and $2$ are competing goals, as larger representative sets allow for lower average distance. In this article we focus primarily on criterion $1$, using criterion $2$ as a secondary evaluation criterion.

\subsection{Testbed Applications}
\label{probmot}
Representative selection is useful in many contexts, particularly when the full dataset is either redundant (due to many near-identical samples) or when using all samples is impractical. For instance, given a large document and a satisfactory measure of similarity between sentences, text summarization \cite{mani1999advances} could be framed as a representative selection task - obtain a subset of sentences that best captures the nature of the document. Similarly, one could map this problem to extracting ``visual words'' or representative frames from visual input \cite{yuan2007discovery,mayol2005wearable}. This paper examines two concrete cases in which representatives are needed:

\begin{itemize}

\item {\bf {\em Music analysis}} - the last decade has seen a rise in the computational analysis of music databases for music information retrieval \cite{casey2008content}, recommender systems \cite{lamere2008social} and computational musicology \cite{cook2004computational}. A problem of interest in these contexts is to extract short representative musical segments that best represent the overall character of the piece (or piece set). This procedure is in many ways analogous to text summarization. 
\item {\bf {\em Team strategy/behavior analysis}} - opponent modeling has been discussed in several contexts, including game playing \cite{billings1998opponent}, real-time agent tracking \cite{tambe1995resc} and general multiagent settings \cite{carmel1996opponent}. Given a large dataset of recorded behaviors, one may benefit from reducing this large set into a smaller collection of prototypes. In the results section, we consider this problem as a second testbed domain.

\end{itemize}

What makes both these domains appropriate as testbeds is that they are realistically rich and induce complex, non-metric distance relations between samples.

The structure of this paper is as follows. In the following section we
provide a more extensive context to the problem of representative
selection and discuss why existing approaches may not be suitable. In
Section $3$ we introduce $\delta$-medoids, an algorithm specifically designed to tackle the problem as we formally defined it. In Section $4$ we show some theoretical analysis of the suggested algorithm, and in Section $5$ we show its empirical performance in the testbed domains described above.

\section{Background and Related Work}

There are several existing classes of algorithms that solve problems
related to representative selection.   This section reviews them and
discusses the extent to which they are (or are not) applicable to our problem.

\subsection{Limitations of Traditional Clustering} 
Given the prevalence of clustering algorithms, it is tempting to solve representative selection by clustering the data and using cluster centers (if they are in the set) as representatives, or the closest point to each center. In some cases it may even seem sufficient, once the data is clustered, to select samples chosen at random from each cluster as representatives. However, such an approach usually only considers the average distance between representatives and samples, and it is unlikely to yield good results with respect to any other requirement, such as minimizing the worst case distance or maintaining the smallest set possible. Moreover, the task of determining the desirable number of clusters $k$ for a sufficient representation can be a difficult challenge in itself.

Consider the example in Figure $1$: given a set of $|S|=100$ points, and a distance measure (in this case, the Euclidean distance metric), we seek a set of representatives that is within distance 1 of every point in the set, thus satisfying Criterion $1$ with $\delta=1$. Applying a standard clustering approach on this set, the distance constraint is only consistently met when $k\geq77$ (and rarely with less than $70$ samples). Intuitively, a large number of clusters is required to ensure that no sample is farther than $\delta$ from a centroid. However, we can obtain the same coverage goal with only $13$ samples. Defining a distance criterion rather than a desired number of clusters has a subtle but crucial impact on the problem definition.
\begin{figure}[h!]
  \label{prefig1}
  \centering
    \includegraphics[width=0.8\textwidth,natwidth=310,natheight=322]{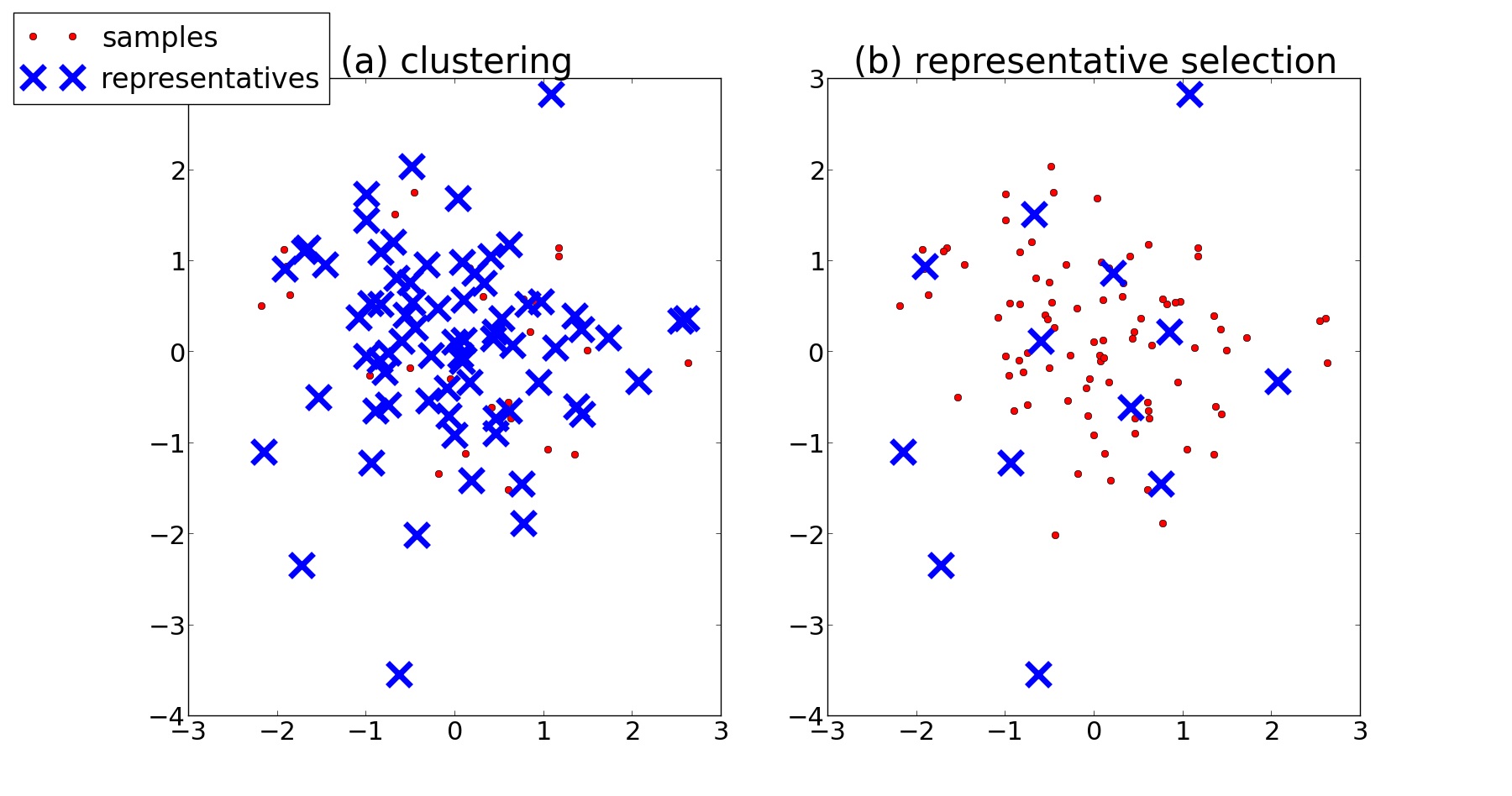}
  \caption{Clustering vs. representative selection. (a) When applying $k$-medoids (defined in Section $2.4$), $k=77$ clusters are required to satisfy the distance condition. (b) A better representative set does so with only $13$ representatives.}
\end{figure}

\subsection{Clustering and Spatial Representation}
The above limitation of clustering applies even when the data can be embedded as coordinates in some $n$-dimensional vector space. However, in many cases, including our motivating domain of music analysis, the data does not naturally fit in such a space. This constraint renders many common clustering techniques inapplicable, including the canonical $k$-means \cite{macqueen1967some}, or more recent works such as \cite{wang2013beyond}. Furthermore, the distance function we construct (detecting both local and global similarities) is not a metric, since it violates the triangle inequality. Because of this property, methods reliant on a distance metric are also inapplicable. Among such methods are neighbor-joining \cite{saitou1987neighbor}, which becomes unreliable when applied on non-metric data, or the $k$-prototypes algorithm \cite{azran2006new}.\footnote{In certain contexts, metric learning \cite{xing2002distance,davis2007information} can be applied, but current methods are not well suited for data without vector space representation, and in some sense, learning a metric is of lesser interest for representative selection, as we care less about classification or the structural relations latent in the data.}

Nevertheless, certain clustering algorithms still apply, such as the $k$-medoids algorithm, hierarchical clustering \cite{sibson1973slink}, and spectral clustering \cite{von2007tutorial}. These methods can be employed directly on a pairwise (symmetric) similarity matrix,\footnote{For spectral clustering, the requirement is actually an affinity (or proximity) matrix.} while satisfying the triangle inequality is not a requirement.

\subsection{The $k$-Medoids Algorithm}
\label{subsec:kmeds}
The $k$-medoids algorithm \cite{rousseeuw1990finding} is a variation on the classic $k$-means algorithm that only selects centers from the original dataset, and is applicable to data organized as a pairwise distance matrix. The algorithm partitions a set of samples to a predetermined number $k$ based on the distance matrix. Similarly to the $k$-means algorithm, it does so by starting with $k$ random centers, partitioning the data around them, and iteratively moving the $k$ centers toward the medoids of each cluster (a medoid is defined as $ medoid_S = \underset{s \in S}{\operatorname{{\it argmin}}} {\sum_{x \in S}{d(x, s)}}$).

All of the approaches mentioned so far are specifically designed for dividing the data to a fixed number of partitions. In contrast, representative selection defines a distance (or coverage) criterion $\delta$, rather than a predetermined number of clusters $k$. In that respect, $k$-medoids, or spectral and hierarchical clustering, force us to search for a partition that satisfies this distance criterion. Applying a clustering algorithm to representative selection requires an outer loop to search for an appropriate $k$, a process which can be quite expensive.

We note that traditionally, both hierarchical methods and spectral clustering require the full pairwise distance matrix. If the sample set $S$ is large (The usual use case for representative selection), computing a pairwise $|S| \times |S|$ distance matrix can be prohibitively expensive. In the case of spectral clustering, an efficient algorithm that does not compute the full distance matrix exists \cite{shu2011efficient}, but it relies on a vector space representation of the data, rendering it inapplicable in our case.\footnote{In some cases the distance matrix can be made sparse via $KD$-trees and nearest-neighbor approximations, which also require a metric embedding.\cite{chen2011parallel}} The algorithm we introduce in this article does not require a distance metric, nor does it rely on such a spatial embedding of the data, which makes it useful even in cases where very little is known about the samples beyond some proximity relation.

\subsection{$k$-Centers Approach}
A different, yet related, topic in clustering and graph theory is the $k$-centers problem. Let the distance between a sample $s$ and a set $C$ be: \linebreak $d(s,C)=min_{c \in C}d(s,c)$. The $k$-centers problem is defined as follows: Given a set $S$ and a number $k$, find a subset $R \subset S, |R|=k$ so that $max_{s \in S} d(s,R)$ is minimal \cite{hochbaum1985best}. 

In metric spaces, an efficient $2$-approximation algorithm for this problem exists as follows.\footnote{We note that no better approximation scheme is possible under standard complexity theoretic assumptions \cite{hochbaum1985best}.} First choose a random representative. Then, for $k-1$ times, add the element farthest away from the representative set $R$ to $R$. This approach can be directly extended to suit  representative selection - instead of repeating the addition step $k-1$ times, we can continue adding elements to the representative set until no sample is $>\delta$ away from any some representative (see Algorithm $1$).

\begin{algorithm}[tb!]
   \caption{Extended Greedy K-Centers Approach (farthest first traversal)}
\label{alg:kcenter}
\begin{algorithmic}[1]
\STATE {\bfseries Input:} data $sampleSet = x_0 \ldots x_m$, required distance $\delta$
\STATE choose random starting representative $x_i$
\STATE representativeSet $= \{x_i\}$
\STATE sampleSet $= x_0 \ldots x_{i-1}, x_{i+1} \ldots x_m$
\STATE maximalDist $= max_{s \in sampleSet}d(s, representativeSet$
\WHILE{maximalDist $> \delta$}
\STATE maximalElement $= argmax_{s \in sampleSet}d(s, representativeSet)$
\STATE representativeSet $= representativeSet \cup \{maximalElement\}$
\STATE sampleSet $= representativeSet / \{maximalElement\}$
\STATE maximalDist $= max_{s \in sampleSet}d(s, representativeSet)$
\ENDWHILE
\end{algorithmic}
\end{algorithm}

While this algorithm produces a legal representative set, it ignores criterion $2$ (average distance).

Another algorithm that is related to this problem is Gonzales' approximation algorithm for minimizing the maximal cluster diameter \cite{gonzalez1985clustering}, which iteratively takes out elements from existing clusters to generate new clusters based on the inter-cluster distance. This algorithm is applicable in our setting since it too only requires pairwise distances, and can produce a legal coverage by partitioning the data into an increasing number of cluster until the maximal diameter is less than $\delta$ (at which point any sample within a cluster covers it). This approach is wasteful for the purpose of representative selection, since it forces a much larger number of representatives than needed.

Lastly, in a recent, strongly related paper \cite{elhamifar2012finding}, the authors consider a similar problem of selecting exemplars in data to speed up learning. They do not pose hard constraints on the maximal distance between exemplars and samples, but rather frame this task as an optimization problem, softly associating each sample with a ``likelihood to represent'' any other sample, and trying to minimize the aggregated coverage distance while also minimizing the norm of the representation likelihood matrix. Though very interesting, it's hard to enable this method to guarantee a desired minimal distance, and the soft association of representatives to samples is inadequate to our purposes. 

\section{The $\delta$-Medoids Algorithm}
\label{sec:dmed}

In this section, we present the novel $\delta$-medoids algorithm, specifically designed to solve the representative selection problem. The algorithm does not assume a metric or a spatial representation, but rather relies solely on the existence of some (not necessarily symmetric) distance or dissimilarity measure $d : S \times S \rightarrow \mathbb{R}^+$. Similarly to the $k$-centers solution approach, the $\delta$-medoids approach seeks to directly find samples that sufficiently cover the full dataset. The algorithm does so by iteratively scanning the dataset and adding representatives if they are sufficiently different from the current set. As it scans, the algorithm associates a cluster with each representative, comprising the samples it represents. Then, the algorithm refines the selected list of representatives, in order to reduce the average coverage distance. This procedure is repeated until the algorithm reaches convergence. Thus, we address both minimality (criterion 1) and average-distance considerations (criterion 2). We show in Section \ref{sec:empres} that this algorithm achieves its goal efficiently in two concrete problem domains, and does so directly, without the need for optimizing a meta-parameter $k$. 

We first introduce a simpler, single-iteration $\delta$-representative selection algorithm on which the full $\delta$-medoids algorithm is based.

\subsection{Straightforward $\delta$-Representative Selection}
Let us consider a more straightforward ``one-shot'' representative selection algorithm that meets the $\delta$-distance criterion. The algorithm sweeps through the elements of $S$, and collects a new representative each time it observes a sufficiently ``new'' element. Such an element needs to be $>\delta$ away from any previously collected representative. The pseudocode for this algorithm is presented in Algorithm $2$.

\begin{algorithm}[tb!]
   \caption{One-shot $\delta$-representatives selection algorithm}
\label{alg:oneshot}
\begin{algorithmic}[1]
   \STATE {\bfseries Input:} data $x_0 \ldots x_m$, required distance $\delta$
   \STATE Initialize $representatives = \emptyset$.
   \STATE Initialize $clusters = \emptyset$
   \STATE {\bf {\small representative assignment subroutine, \emph{RepAssign}, lines 5-22:}}
   \FOR{$i=0$ {\bfseries to} $m$}
   \STATE Initialize $dist = \infty$
   \STATE Initialize $representative = null$
   \FOR{$rep$ in $representatives$}
   \IF {$d(x_i, rep) \leq dist$}
   \STATE $representative = rep$
   \STATE $dist = d(x_i, rep)$
   \ENDIF
   \ENDFOR
   \IF{$dist \leq \delta$}
   \STATE add $x_i$ to $cluster_{representative}$
	\ELSE
   \STATE $representative = x_i$
   \STATE Initialize $cluster_{representative} = \emptyset$
   \STATE add $x_i$ to $cluster_{representative}$
   \STATE add $cluster_{representative}$ to $clusters$
   \ENDIF

   \ENDFOR
\end{algorithmic}
\end{algorithm}

While this straightforward approach works well in the sense that it does produce a legal representative set, it is sensitive to scan order, therefore violating the desired stability property. More importantly, it does not address the average distance criterion. For these reasons, we extend this algorithm into an iterative one, a hybrid of sorts between direct representative selection and $EM$ clustering approaches.

\subsection{The Full $\delta$-Medoids Algorithm}

This algorithm is based on the straightforward approach, as described in Section 3.1. However, unlike Algorithm $2$, it repeatedly iterates through the samples. In each iteration, the algorithm associates each sample to a representative so that it is never $\geq \delta$ away from some representative (the \emph{RepAssign} subroutine, see Algorithm $3$), just as in Algorithm $2$. The main difference is that at the end of each iteration it subsequently finds a closer-fitting representative for each cluster $S$ associated with representative $s$. Concretely, $ representative_S = medoid_S = \underset{s \in S}{\operatorname{{\it argmin}}} {\sum_{x \in S}{d(x, s)}}$ (lines $8-13$), under the constraint that \emph{no sample $\in S$ is farther than $\delta$ from $medoid_S$}. This step ensures that a representative is ``best-fit'' on average to the cluster of samples it represents, without sacrificing coverage. In other words, while trying to minimize the size of the representative set, the algorithm also addresses criterion $2$ - average distance as low as possible. This step also drastically improves the stability of the retrieved representative set under different permutations of the data (desideratum $1$). We note that by adding the constraint that new representatives must still cover the clusters they were selected from, we guarantee that the number of representatives $k$ does not increase after the first scan. 

The process is repeated until $\delta$-medoids reaches convergence, or until we reach a representative set which is ``good enough'' (remember that at the end of each cluster-association iteration we have a set that satisfies the distance criterion). This algorithm uses a greedy heuristic that is indeed ensured to converge to some local optimum (Theorem \ref{dmedoidconv}). This local optimum is dependent on the value of $\delta$ and the structure of the data. In Subsection \ref{subsec:hardness}, we show that solving the representative selection problem for a given $\delta$ is NP-hard, and therefore heuristics are required.

\begin{algorithm}[tb!]
   \caption{The $\delta$-medoid representative selection algorithm.}
\label{alg:dmedoid}
{\fontsize{7}{7}\selectfont
\begin{algorithmic}[1]
   \STATE {\bfseries Input:} data $x_0 \ldots x_m$, required distance $\delta$
   \STATE $t = 0$
   \STATE Initialize $representatives_{t=0} = \emptyset$.
   \STATE Initialize $clusters = \emptyset$
   
   \REPEAT
   \STATE $t = t + 1$

   \STATE {\bf call \emph{RepAssign} subroutine, lines 5-22 of Algorithm 2}

   \STATE Initialize $representatives_t = \emptyset$
   \FOR{$cluster$ in $clusters$}
   \STATE $representative = \underset{s \in cluster}{\operatorname{{\it argmin}}} {\sum_{x \in cluster}{d(x, s)}}$ s.t.  $\forall x \in cluster. d(x,s) \leq \delta$
   \STATE add $representative$ to $representatives_t$
   \ENDFOR
   \UNTIL{$representatives_t \equiv representatives_{t-1}$}
\end{algorithmic}}

\end{algorithm}

\begin{mydef}
\label{dmedoidconv}
Algorithm $3$ converges after a finite number of steps.
\end{mydef}
See Appendix A for proof sketch.

\subsection{Merging Close Clusters}

Since satisfying the distance constraint with a minimal set of representatives (Criterion $1$) is NP-hard (see Section $4$), the $\delta$-medoids algorithm is not guaranteed to do so. A simple optimization procedure can reduce the number of representatives in certain cases.
For instance, in some cases, oversegmentation may ensue. To abate such an occurrence, it is possible to iterate through representative pairs that are no more than $\delta$ apart, and see whether joining their respective clusters could yield a new representative that covers all the samples in the joined clusters. If it is possible, the two representatives are eliminated in favor of the new joint representative. The process is repeated until no pair in the potential pair list can be merged. This procedure can be generalized for larger representative group sizes, depending on computational tractability. These refinement steps can be taken after each iteration of the algorithm. If the number of representatives is high, however, this approach may be computationally infeasible altogether. Although this procedure was not required in our problem domains (see Section \ref{sec:empres}), we believe it may still prove useful in certain cases.

\section{Analysis Summary}
In this section, we present the hardness of the representative selection problem, and briefly discuss the efficiency of the $\delta$-medoids algorithm. We show that the problem of finding a minimal representative set is NP-hard, and provide certain bounds on the performance of $\delta$-medoids in metric spaces with respect to representative set size and average distance. We continue to show that approximating the representative selection problem is NP-hard in non-metric spaces, both in terms of the representative set size and with respect to the maximal distance.

For the sake of readability, we present full details in Appendix D.

\subsection{NP-Hardness of the Representative Selection Problem}
\label{subsec:hardness}

\begin{mydef}
Satisfying Criterion $1$ (minimal representative set) is NP-Hard.
\end{mydef}

\subsection{Bounds on $\delta$-medoids in Metric Spaces}

The $\delta$-medoids algorithm is agnostic to the existence of metric space in which the samples can be embedded. That being said, it can work equally well in cases where the data is metric (in Appendix C we demonstrate the performance of the $\delta$-medoids algorithm in a metric space test-case). However, we can show that if the measure which generates the pairwise distances is in fact a metric, certain bounds on performance exist.

\begin{mydef}
\label{twoapprox}
In a metric space, the average distance of a representative set $|C|=k$ obtained by the $\delta$-medoids algorithm is bound by $2OPT$ where $OPT$ is the maximal distance obtained by an optimal assignment of $k$ representatives (with respect to maximal distance).
\end{mydef}

\begin{mydef}
The size of the representative set returned by the $\delta$-medoids algorithm, $k$, is bound by $k \leq N(\frac{\delta}{2})$ where $N(x)$ is the minimal number of representatives required to satisfy distance criterion $x$.
\label{corollar}
\end{mydef}

\subsection{Hardness of Approximation of Representative Selection in Non-Metric Spaces}

In non-metric spaces, the representative selection problem becomes much harder. We now show that no $c$-approximation exists for the representative selection problem either with respect to the first criterion (representative set size) or the second criterion (distance - we focus on maximal distance but a similar outcome for average distance is implied).

\begin{mydef}
No constant-factor approximation exists for the representative selection set problem with respect to representative set size.
\end{mydef}

\begin{mydef}
For representative sets of optimal size $k$,\footnote{In fact, this proof applies for any value of $k$ that cannot be directly manipulated by the algorithm.} no constant-factor approximation exists with respect to the maximal distance between the optimal representative set and the samples.
\end{mydef}

\subsection{Efficiency of $\delta$-Medoids}

The actual run time of the algorithm is largely dependent on the data and the choice of $\delta$. An important observation is that at each iteration, each sample is only compared to the current representative set, and a sample is introduced to the representative set only if it is $>\delta$ away from all other representatives. After each iteration, the representatives induce a partition to clusters and only samples within the same cluster are compared to one another. While in the worst case the runtime complexity of the algorithm can be $O(|S|^2)$, in practice we can get considerably better runtime performance, closer asymptotically to $|S|^{1.5}$. We note that in each iteration of the algorithm, after the partitioning phase (the \emph{RepAssign} subroutine in Algorithm \ref{alg:dmedoid}) the algorithm maintains a legal representative set, so in practice we can halt the algorithm well before convergence, depending on need and resources.

\section{Empirical Results}
\label{sec:empres}

In this section, we analyze the performance of the $\delta$-medoids algorithm empirically in two problem domains - music analysis and agent movement analysis. We show that $\delta$-medoids does well on Criterion $1$ (minimizing the representative set) while obtaining a good solution for Criterion $2$ (maintaining a low average distance). We compare ourselves to three alternative methods - $k$-medoids, the greedy $k$-center heuristic, and spectral clustering (using cluster medoids as representatives) \cite{shi2000normalized}, and show we outperform all three. We note that although these methods weren't necessarily designed to tackle the representative selection problem, they, and clustering approaches in general, are used for such purposes in practice (see \cite{hadi2006video}, for instance). To obtain some measure of statistical significance, for each dataset we analyze, we take a random subset of $|S|=5000$ samples and use this subset as input for the representative selection algorithm. We repeat this process $N=20$ times, averaging the results and obtaining standard errors. We show that the $\delta$-medoid algorithm produces representative sets at least as compact as those produced by the $k$-centers approach, but obtains a much lower average distance. We further note it does so directly, without the need for first optimizing the number of clusters $k$, unlike $k$-medoids or spectral clustering.

In Appendix C, we also demonstrate the performance of the algorithm in a standard metric space, and show it outperforms the other methods in this setting as well.

\subsection{Distance Measures}

In both our problem domains, no simple or commonly accepted measure of distance between samples exists. For this reason, we devised a distance function for each setting, based on domain knowledge and experimentation. Feature selection and distance measure optimization are beyond the scope of this work. For completeness, the full details of our distance functions appear in Appendix E. We believe the results are not particularly sensitive to the choice of a specific distance function, but we leave such analysis to future work.

\subsection{Setting 1 - musical segments}

In this setting, we wish to summarize a set of musical pieces. This domain illustrates many of the motivations listed in Section $1$. The need for good representative selection is driven by several tasks, including style characterization, comparing different musical corpora (see \cite{dubnov2003using}), and music classification by composer, genre or period \cite{bergstra2006aggregate}.

For the purpose of this work we used the Music21 corpus, provided in MusicXML format \cite{cuthbert2010music21}. For simplicity, we focus on the melodic content of the piece, which can be characterized as the variation of pitch (or frequency) over time.

\subsubsection{Data}

We use thirty musical pieces: 10 representative pieces by Mozart, Beethoven, and Haydn. The melodic lines in the pieces are isolated and segmented using basic grouping principles adapted from \cite{pearce2008comparison}. In the segmentation process, short overlapping melodic sequences $5$ to $8$ beats long are generated.  For example, three such segments are plotted in Figure $2$ as pitch variation over time. For each movement and each instrument, segmentation results in $55-518$ segments. All in all we obtain $20,000-40,000$ segments per composer.

\begin{figure}[h!]
  \label{fig2}
  \centering
    \includegraphics[width=\textwidth]{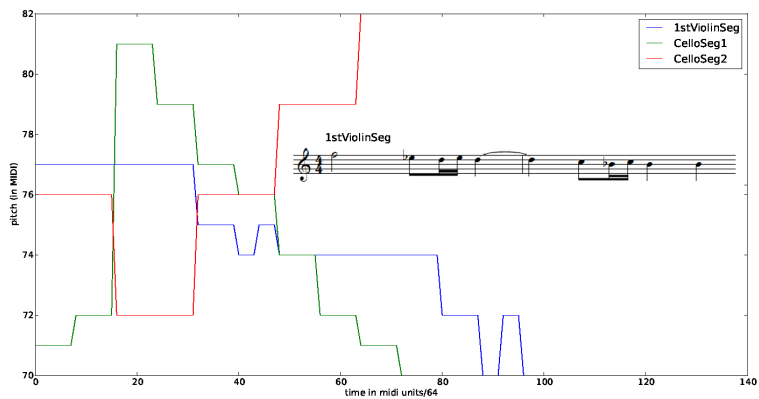}
  \caption{Three musical segments as pitch (in MIDI format) over time, along with the musical notation of the first segment (1stViolinSeg).}
\end{figure}

\subsubsection{Distance measure}

We devise a fairly complex distance measure between any two musical segments $S_1$ and $S_2$. Several factors are taken into account:

\begin{itemize}
\item {\em Global alignment} - the global alignment score between the two segments, calculated using the Needleman-Wunsch algorithm \cite{needleman1970general}.
\item {\em Local alignment} - the local alignment score between the two segments,  calculated using the Smith-Waterman algorithm \cite{SmithWaterman}. Local alignment is useful if two sequences are different overall, but share a meaningful subsequence.
\item {\em Rhythmic overlap, interval overlap, step overlap, pitch overlap} - the extent to which one-step melodic and rhythmic patterns in the two segments overlap, using a ``bag''-like distance function $d_{set}(A_1, A_2)=\frac{|A_1 \bigtriangleup A_2|}{|A_1 \cup A_2|}$.
\end{itemize}

The different factors are then weighted and combined. This measure was chosen because similarity between sequences is multifaceted, and the different factors above capture different aspects of similarity, such as sharing a general contour (global alignment), a common motif (local alignment), or a similar ``musical vocabulary'' (the other factors, which by themselves each capture a different aspect of musical language). The result is a measure but not a metric since local alignment may violate the triangle inequality.

\subsubsection{Results}
We compare $\delta$-medoids to the $k$-medoids algorithm and the greedy $k$-center heuristic for five different $\delta$ values. The results are presented in Figure $3$. For each composer and $\delta$, we searched exhaustively for the lowest $k$ value for which $k$-medoids met the distance requirement. We study both the size of the representative set obtained and the average sample-representative distance.
\begin{figure}[htb]
  \label{fig3}
  \centering
    \includegraphics[width=\textwidth,natwidth=310,natheight=322]{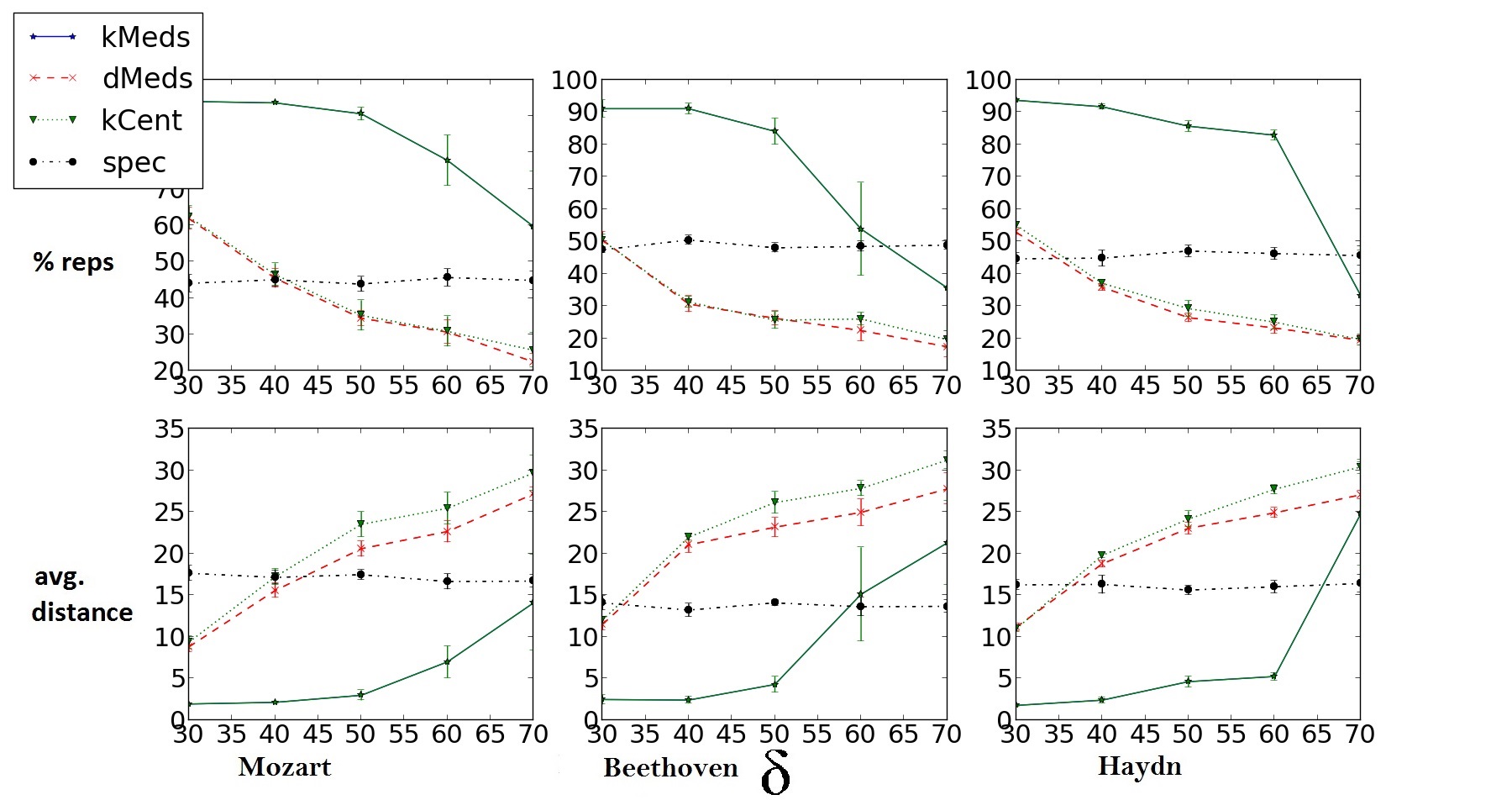}
  \caption{Representative set size in \% from entire set and average representative set distance for three different composers, ten different pieces each, and five different distance criteria. Each column represents data for a different composer. $\delta$-medoids yields the most compact representative set overall while still obtaining a smaller average distance than the $k$-centers heuristic.}
\end{figure}

From the representative set size perspective, for all choices of $\delta$ the $\delta$-medoids algorithm obtains better coverage of the data compared to the $k$-medoids, and does at least as well (and most often better) compared to the greedy $k$-centers heuristic. However, in terms of average distance, $\delta$-medoids performs much better compared to the $k$-centers heuristic, implying that the $\delta$-medoids algorithm outperforms the other two. While spectral clustering seems to satisfy the distance criteria with a small representative set for small values of $\delta$, its non-centroid based nature makes it less suitable for representative selection, as a more lax $\delta$ criterion might not necessarily mean a smaller representative set will be needed (as apparent from the result). Indeed, as the value of $\delta$ increases, the $\delta$-medoids algorithm significantly outperforms spectral clustering.

\subsection{Setting 2 - agent movement in robot soccer simulation}
As described in Section \ref{probmot}, analyzing agent behavior can be of interest in several domains. The robot world-cup soccer competition (RoboCup) is a well-established problem domain for AI in general \cite{kitano1997robocup}. In this work we chose to focus on the RoboCup 2D Simulation League. We have collected game data from several full games from the past two Robocup competitions.
An example for the gameplay setting and potential movement trajectories can be seen in Figure $5$.

\begin{figure}[!h]
  \label{fig4}
  \centering
    \includegraphics[width=\textwidth,natwidth=610,natheight=642]{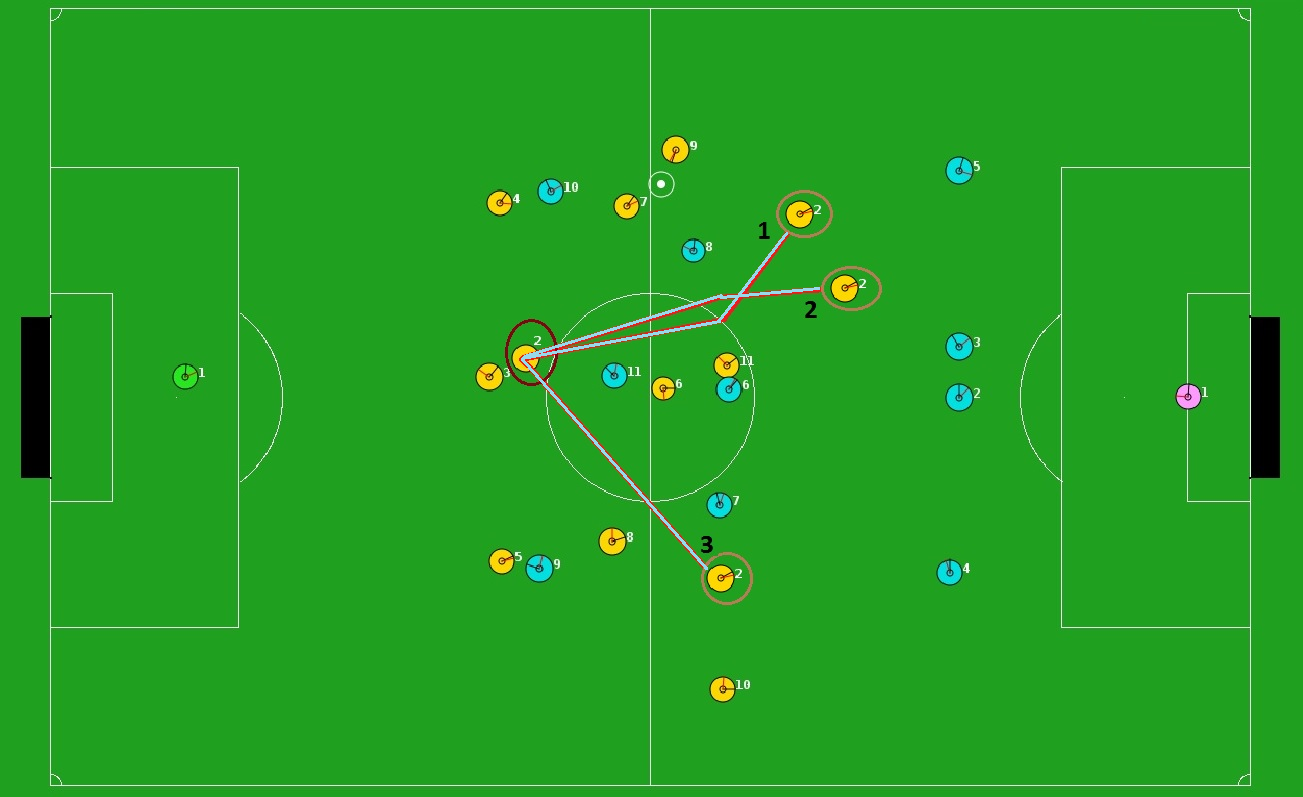}
      \caption{The RoboCup 2D Simulation. Three potential movement trajectories for a specific agents are marked.}
\end{figure}

Our purpose is to extract segments that best represent agent movement patterns throughout gameplay. In the specific context of the Robocup simulation league, there are several tasks that motivate representative selection, including agent and team characterization, and learning training trajectories for optimization.

\subsubsection{Data}

Using simulation log data, we extract the movement of the $22$ agents over the course of the game (\#$timesteps = 6000$). The agents move in $2$-dimensional space (three example trajectories can be seen in Figure $5$). We extract $1$-second ($10$ timestamps) long,  partially overlapping segments from the full game trajectories of all the agents on the field except the goalkeeper, who tends to move less and in a more confined space and for the purpose of this task is of lesser interest. That leads to $900 \cdot 20 = 18000$ movement segments in total per game. We analyzed $4$ teams and $5$ games ($90000$ segments) per team.

\subsubsection{Distance measure}

Given two trajectories, one can compare them as contours in 2-dimensional space. We take an alignment-based approach, with edit costs being the RMS distance between them. Our distance measure is comprised of three elements: global and local alignment (same as in music analysis), and a ``bag of words''-style distance based on patterns of movement-and-turn sequences (turning is quantized into $6$ angle bins). As in music analysis, the reason for this approach is that similarity in motion is hard to define, and we believe each feature captures different aspects of similarity. As in the previous setting, this is not a metric, as local alignment may violate the triangle inequality.

\subsubsection{Results}

As in the previous setting, we compare $\delta$-medoids to the $k$-medoids algorithm as well as the greedy $k$-center heuristic, for five different game logs and five different $\delta$ values. The results are presented in Figure $5$. As before, for each $\delta$, we searched exhaustively for the optimal choice of $k$ in $k$-medoids.
\begin{figure}[htb]
  \label{fig5}
  \centering
    \includegraphics[width=\textwidth,natwidth=310,natheight=322]{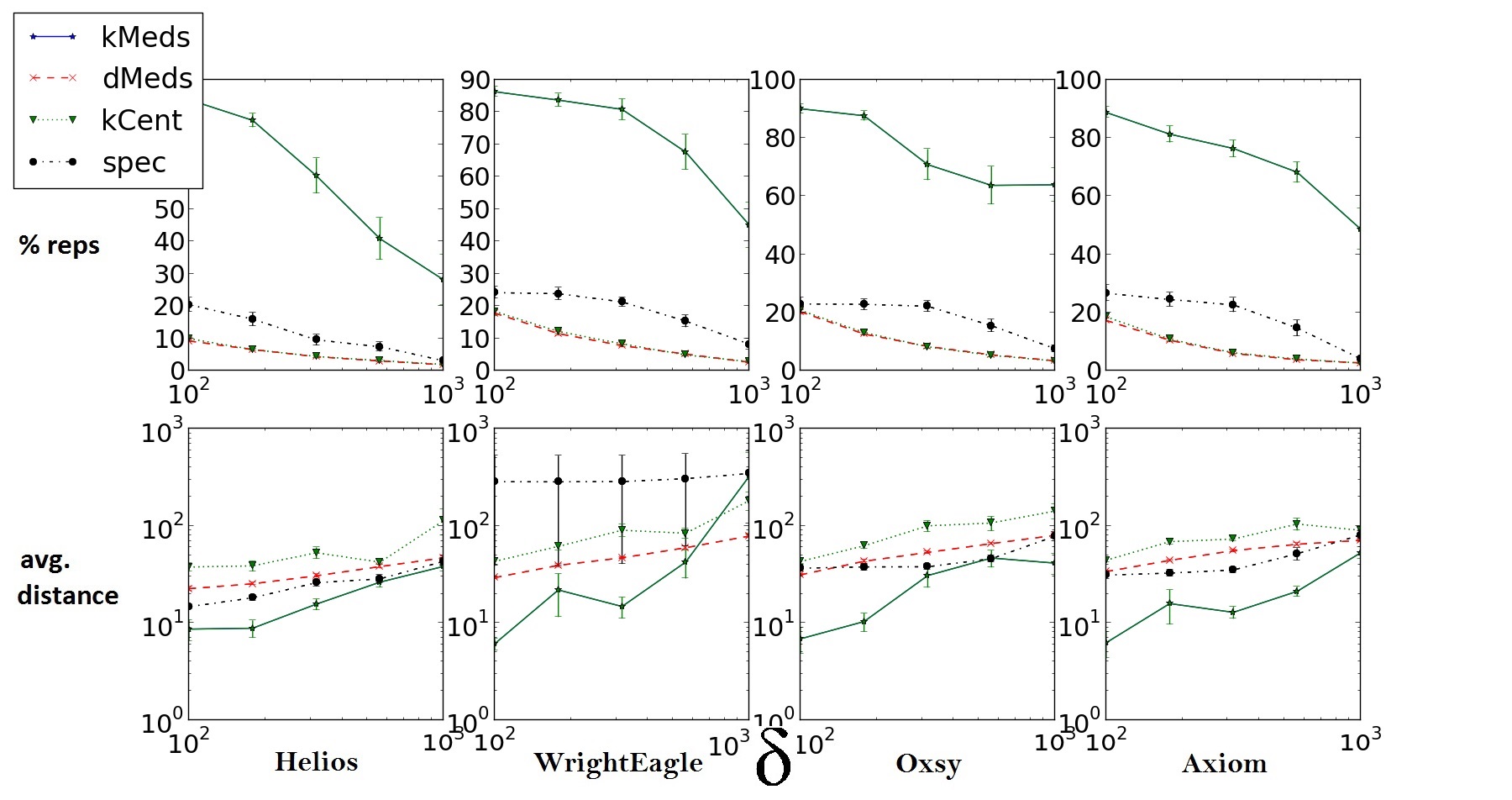}
  \caption{Representative set size in \% from entire set for four different teams, five different game logs each, and five distance criteria. Each column represents game data for a different team. Axes denoting distance are in log-scale.}
\end{figure}

The results reinforce the conclusions reached in the previous domain - for all choices of $\delta$ we meet the distance requirement using a much smaller representative set compared to the $k$-medoids and spectral clustering approaches (which does much worse in this domain compared to the previous one). Furthermore, $\delta$-medoids once again does at least as well as the $k$-centers heuristic. In terms of average distance, our algorithm performs much better compared to the $k$-centers heuristic, suggesting that the $\delta$-medoids algorithm generally outperforms the other approaches.

\subsection{Stability of the $\delta$-Medoids Algorithm}

In this section we establish that indeed the $\delta$-medoids algorithm is robust with respect to scan order (satisfying desideratum $1$ from Section $1$). To test stability, we ran $\delta$-medoids, $k$-medoids, the $k$-center heuristic and spectral clustering multiple times on a large collection of datasets, reshuffling the input order on each iteration, and examined how well preserved the representative set was across iterations and methods. Our analysis indicated that the first three algorithms consistently obtain $>90$\% average overlap, and the level of stability observed is almost identical. Spectral clustering yields drastically less stable representative sets. For a fuller description of these results see Appendix B.

%

\section{Summary and Discussion}	

In this paper, we present a novel heuristic algorithm to solve the representative selection problem: finding the smallest possible representative subset that best fits the data under the constraint that no sample in the data is more than a predetermined parameter $\delta$ away from some representative. We introduce the novel $\delta$-medoids algorithm and show it outperforms other approaches that are only concerned with either best fitting the data into a given number of clusters, or minimizing the maximal distance.

There is a subtle yet significant impact to focusing on a maximal distance criterion $\delta$ rather than choosing the number of clusters $k$. While both $\delta$-medoids and $k$-medoids aim to minimize the sum of distances between representatives and the full set, $k$-medoids does so with no regard to any individual distance. Because of this, we need to increase the value of $k$ drastically in order to guarantee that our distance criterion is met, and that sparse regions of our sample set are sufficiently represented. This results in over-representation of dense regions in our sample set. By carefully balancing between minimality under the distance constraint and average distance minimization, the $\delta$-medoids algorithm adjusts the representation density adaptively based on the sample set, without any prior assumptions.

Although this paper establishes $\delta$-medoids as a leading algorithm for representative selection, we believe that more sophisticated algorithms can be developed to handle different variations of this problem, putting different emphasis on the minimality requirement for the representative set vs. how well the set fits the data. Depending on the specific nature of the task the representatives are needed for, different tradeoffs may be most appropriate and lead to algorithmic variations. For instance, an extension of interest could be to modify the value of $\delta$ adaptively depending of the density of sample neighborhoods. However, we show that $\delta$-medoids is a promising approach to the general problem of efficient representative selection.

\begin{small}
\subsection*{Acknowledgements}
This work has taken place in the Learning Agents Research
Group (LARG) at the Artificial Intelligence Laboratory, The University
of Texas at Austin.  LARG research is supported in part by grants from
the National Science Foundation (CNS-1330072, CNS-1305287), ONR
(21C184-01), AFOSR (FA8750-14-1-0070, FA9550-14-1-0087), and Yujin Robot.

\end{small}

\bibliography{refs}
\bibliographystyle{plain}

\appendix

\section{Proof of Convergence for the $\delta$-medoids Algorithm}
In this section, we show that the full proof that the $\delta$-medoids algorithm converges in finite time. 

\begin{mydef2}
Algorithm $3$ converges after a finite number of steps.
\end{mydef2}
\begin{proofsketch}
For any sample $s$ let us denote its associated cluster representative at iteration $i$ $C_i(s)$. Let us denote the distance between the sample and its associated cluster representative as $d(s, C_i(s))$. Observe the overall sum of distances from each point to its associated cluster representative, $\sum_s{d(s, C_i(s))}$.
Assume that after the $i$-th round, we obtain a partition to $k$ clusters, $C_1 .. C_k$. Our next step is to go over each cluster and reassign a representative sample that minimizes the sum of distances from each point to the representative of that cluster, $ \underset{s \in S}{\operatorname{{\it argmin}}} {\sum_{x \in S}{d(x, s)}}$, under the constraints that all samples within the cluster are still within $\delta$ distance of the representative. Since this condition holds prior to the minimization phase, the new representative must still either preserve or reduce the sum of distances within the cluster.

We do this independently for each cluster. If the representatives are unchanged, we have reached convergence and the algorithm stops. Otherwise, the overall sum of distances is diminished. At the $(i+1)$-ith round, we reassign clusters for the samples. A sample can either remain within the same cluster or move to a different cluster. If a sample remains in the same cluster its distance from its associated representative is unchanged. On the other hand, if it moves to a different cluster it means that $d(s, C_i(s)) > d(s, C_{i+1}(s))$, necessarily, so the overall sum of distances from associated cluster representatives is reduced. Therefore, after each iteration we either reach convergence or the sum of distances is reduced. Since there is a finite number of samples, there is a finite number of distance sums, which implies that the algorithm must converge after a finite number of iterations.
\end{proofsketch}
\setcounter{subsection}{0}

\section{Stability of $\delta$-Medoids}

In this section we establish that indeed the $\delta$-medoids algorithm is robust with respect to scan order (satisfying desideratum $1$ from Section $1$). To test this issue, we generated a large collection ($N=1000$) of datasets (randomly sampled from randomly generated multimodal distributions). For each dataset in the collection, we ran the algorithm \#$repetitions=100$ times, each time reshuffling the input order. Next, we calculated the average overlap between any two representative sets generated by this procedure for the same dataset. We then calculated a histogram of average overlap score over all the data inputs. Finally, we compared these stability results to those obtained by the $k$-medoids algorithm (which randomizes starting positions), the $k$-centers heuristic (which randomizes the starting node), and spectral clustering (which uses $k$-means to partition the eigenvectors of the normalized Laplacian). Our analysis indicated that for the first three algorithms, in more than $90$\% of the generated datasets, there was a $>90$\% average overlap. The overlaps observed are almost exactly the same, implying the expected extent of overlap is dependent more on the structure of the data than on the type of randomization the algorithm employs. This serves as evidence that $\delta$-medoids is sufficiently stable, as desired. It should be noted that spectral clustering yields drastically less stable results (ranging between $15$\% and $40$\% overlap), implying a heightened level of stochasticity in the partitioning phase. A histogram indicating our results can be found in Figure $6$. 
\begin{figure}[htb]

 \begin{center}
    \includegraphics[width=\textwidth,natwidth=310,natheight=322]{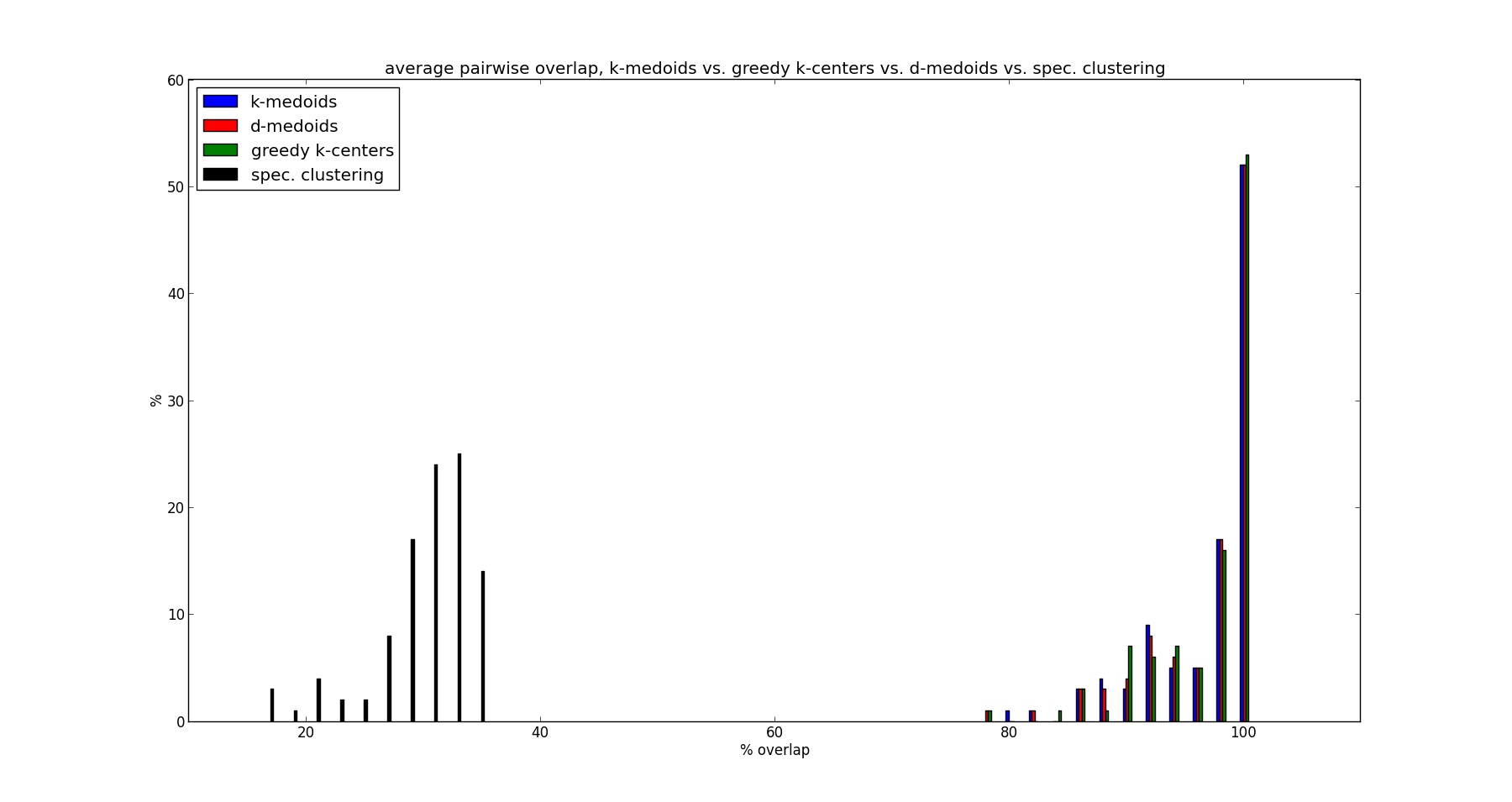}
  \caption{{\small The histograms (plotted as density functions, i.e. counts normalized as percentages) of the average overlap between representative sets found for each method for the same data under different permutations (overlap measured in \%). For $k$-medoids, $\delta$-medoids and the $k$-centers heuristic, in more than 90\% of the datasets, there was a $>90$\% average overlap. Spectral clustering yields drastically less consistent representation sets. The overlaps observed are almost exactly the same, implying the expected extent of overlap is dependent more on the structure of the data than on the type of randomization the algorithm employs.}}
  \end{center}
\end{figure}
One can see that our algorithm has virtually identical stability compared to both $k$-medoids and the greedy $k$-center approaches (which, as stated before, are not sensitive to scan order but contain other types of randomization).

\section{Performance of $\delta$-Medoids in Metric Spaces}
As we state in the paper, though the $\delta$-medoids algorithm is designed to handle non-metric settings, it can easily be used in metric cases as well. In this section we compare the performance of the algorithm to the benchmark methods used in the Empirical Results section. To generate a standard metric setting, we consider a 10-dimensional metric space where samples are drawn from a multivariate Gaussian distribution. We sample a $1000$ samples per experiment, $20$ experiments per setting, with randomly chosen means and variances. The results are presented in Figure $7$.

\begin{figure}[htb]
  \label{fig55}
  \centering
    \includegraphics[width=\textwidth,natwidth=310,natheight=322]{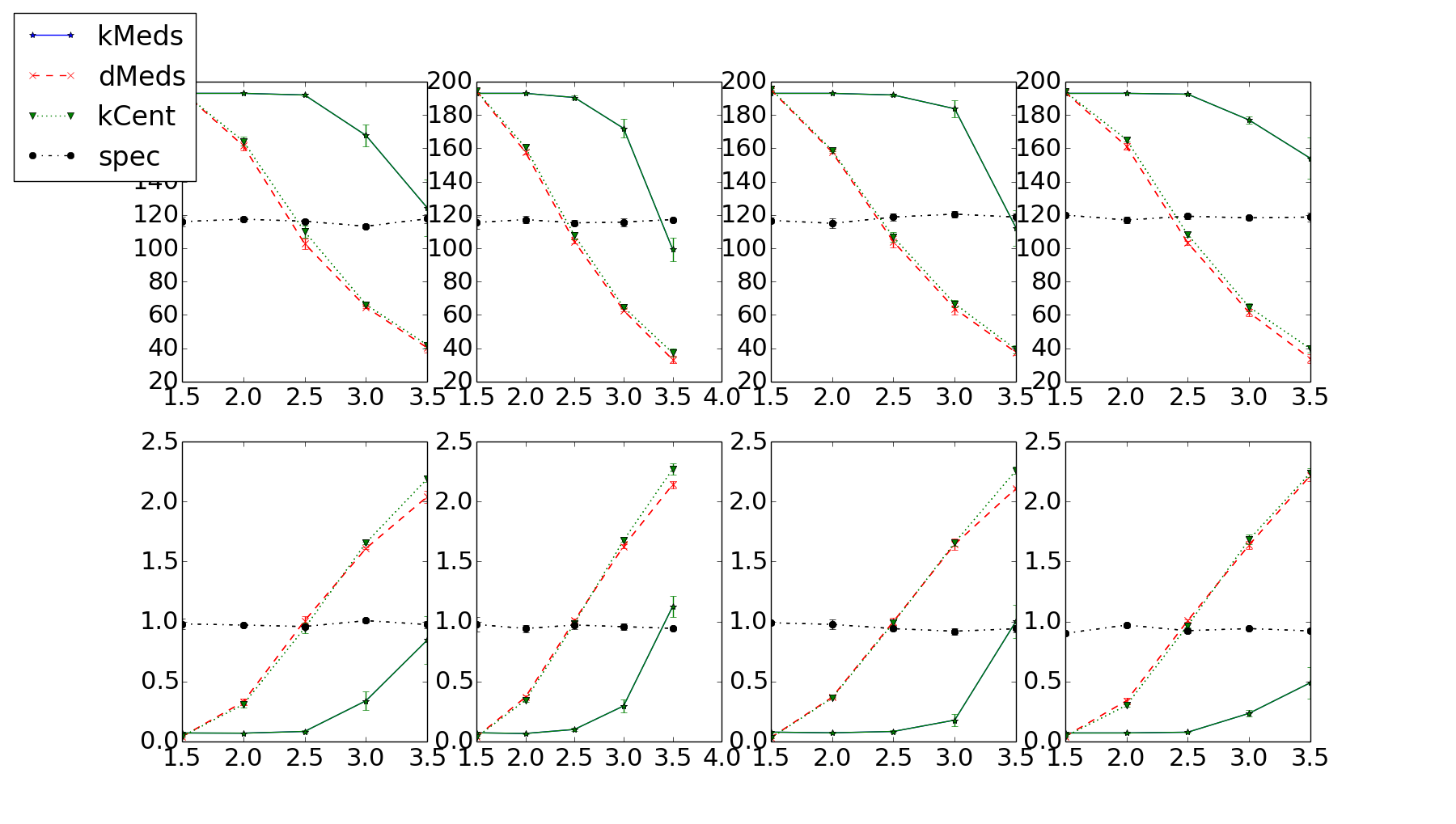}
  \caption{Representative set size in \% from entire set and average representative set distance for four different multivariate Gaussian distributions from which the samples are drawn, $20$ different experiments each, and four different distribution values. Each column represents data for a different distribution. $\delta$-medoids yields the most compact representative set overall while still obtaining a smaller average distance than the $k$-centers heuristic.}
\end{figure}

As one observe, the performance of the $\delta$-medoids algorithm relative to the other methods is qualitatively the same compared to the non-metric cases, despite the metric property of the data in this setting.

\section{Extended Analysis}
In this section, we consider the hardness of the representative selection problem, and discuss the efficiency of the $\delta$-medoids algorithm.

\subsection{NP-Hardness of the Representative Selection Problem}
\label{subsec:hardness2}

\begin{mydef2}
Satisfying Criterion $1$ (minimal representative set) is NP-Hard.
\end{mydef2}

\begin{proofsketch}
We show this via a reduction from the vertex cover problem. Given a graph $G=(V,E)$ we construct a distance matrix $M$ of size $|V| \times |V|$. If two different vertices in the graph, $v_i$, $v_j$, $i \neq j$, are connected, we set the value of entries $(i,j)$ and $(j,i)$ in $M$ to be $\delta-1$. Otherwise, we set the value of the entry to $\delta+1$. Formally: $$M(i,j) = \begin{cases} \delta-1 &\mbox{if } (i, j) \in E \\
0 & \mbox{i = j} \\ \delta+1 & \mbox{otherwise} \end{cases}$$ 
This construction is polynomial in $|V|$ and $|E|$. Let us assume we know how to obtain the optimal representative set for $\delta$ in this case. Then the representative set $S_{rep}$ can be easily translated back to a vertex cover for graph $G$ - simply choose all the vertices that correspond to members of the representative set. Every sample $i$ in the sample set induced by $M$ has to be within $\delta$ range of some representative $j$, meaning that there is an equivalent edge $(i,j) \in E$, which $j$ covers. Since the representative set is minimal, the vertex set is also guaranteed to be the minimal. Therefore, if we could solve the representative selection problem efficiently, we could also solve the vertex cover problem. Since vertex cover is known to be NP-hard \cite{karp}, then so is representative selection.
\end{proofsketch}

\subsection{Bounds on $\delta$-medoids in Metric Spaces}

The $\delta$-medoids algorithm is agnostic to the existence of metric space in which the samples can be embedded. However, we can show that given that the distance measure which generates the pairwise distances is in fact a metric, certain bounds on performance ensue.

\begin{mydef2}
\label{twoapprox2}
In a metric space, the average distance of a representative set $|C|=k$ obtained by the $\delta$-medoids algorithm is bound by $2OPT$ where $OPT$ is the maximal distance obtained by an optimal assignment of $k$ representatives (with respect to maximal distance).
\end{mydef2}

To prove this theorem, and the following one, we will first prove the following helper lemma:

\begin{lemmi}
\label{myLem}
In a metric space, the maximal distance of a representative set $|C|=k$ obtained by the one-shot $\delta$-representative algorithm (Algorithm $2$) is bound by $2OPT$ where $OPT$ is the maximal distance obtained by an optimal assignment of $k$ representatives (with respect to maximal distance).
\end{lemmi}

\begin{proofsketch}
Let $|K|=k$ be the representatives set returned by Algorithm $2$. Let $a^*$ be the sample which is the farthest of any points in the representative set, and let that distance be $\delta^*$. Consider the set $K \cup \{a^*\}$. All $k+1$ points in this set must be of distance $>\delta^*$ from one another - the algorithm would not select representatives of distance $\leq \delta$ from one another, and $\delta \geq \delta^*$, whereas $a^*$ is defined as being exactly $\delta^*$ away from any point in $K$. Let us consider the optimal assignment of $k$ representatives, $K^*$, and let $OPT$ be the maximal distance it achieves. By the pigeonhole principle, at least two samples in the set $K \cup \{a^*\}$ must be associated with the same representative. W.log, let us call these samples $x_1$ and $x_2$, and $k^* \in K^*$ their associated representative. Since the distance between $x_1$ and $x_2$ is greater than $\delta^*$, and since this is a metric space, by the triangle inequality, the distance of $k^*$ from either cannot be smaller than $\frac{\delta^*}{2}$. Therefore $\delta^* < 2OPT$.
\end{proofsketch}

\vspace{2mm}
This implies that algorithm $2$ is asymptotically equivalent to the $k$-centers farthest-first traversal heuristic with respect to maximal distance.
\vspace{2mm}

Now we can prove Theorem \ref{twoapprox2}.

\begin{proofsketch}

First, let us consider Algorithm $2$ (one-shot $\delta$-representatives) on which the $\delta$-medoids algorithm is based. By Lemma \ref{myLem}, the maximal distance obtained by it for a representative set of size $k$ is $<2OPT$, where $OPT$ is the maximal distance obtained by an optimal solution of size $k$ (with respect to maximal distance). The average distance obtained by Algorithm $2$ cannot be greater than the maximal distance, so the same bound holds  the average distance as well. Now let us consider the full $\delta$-medoids algorithm - by definition, it can only reduce the average distance (while maintaining the same representative set size). So the average distance obtained by the $\delta$-medoids algorithm must be bound by $2OPT$ as well.
\end{proofsketch}

\begin{mydef2}
The size of the representative set returned by the $\delta$-medoids algorithm, $k$, is bound by $k \leq N(\frac{\delta}{2})$ where $N(x)$ is the minimal number of representatives required to satisfy distance criterion $x$.
\label{corollary2}
\end{mydef2}

\begin{proofsketch}
By Lemma \ref{myLem}, the maximal distance obtained by it for a representative set of size $k$ is $<2OPT$, where $OPT$ is the maximal distance obtained by an optimal solution of size $k$ (with respect to maximal distance). Let $N(\delta)$ be the covering number for the sample set and distance criterion $\delta$ - that is, the smallest number of representative required so that no sample is farther than $\delta$ from a representative. The size of the representative set returned by the $\delta$-medoids algorithm, is bound by $k \leq N(\frac{\delta}{2})$. Since the full $\delta$-medoids algorithm (Algorithm $3$) first runs Algorithm $2$ and is guaranteed to never increase the size of the representative set size, the same bound holds for it as well.
\end{proofsketch}

In $R^d$, the covering number $N(\epsilon)$ is bound by $O(\frac{1}{\epsilon^d})$. Given that $N(\delta) \geq K^*$ where $K^*$ is the optimal selection of representatives, this implies the solution returned by $\delta$-medoids is bound by a factor of $O(2^d)$. 

It is equivalent to the similar bound known for the $k$-center heuristic \cite{hochbaum1985best,hochbaum1986unified}.

\subsection{Hardness of Approximation of Representative Selection in Non-Metric Spaces}

In non-metric spaces, the representative selection problem becomes much harder. We now show that no $c$-approximation exists for the representative selection problem either with respect to the first criterion (representative set size) or the second criterion (distance - we focus on maximal distance but a similar outcome for average distance is implied).

\begin{mydef2}
No constant-factor approximation exists for the representative selection set problem with respect to representative set size.
\end{mydef2}

\begin{proofsketch}
We show this via a reduction from the set cover problem. Given a set of $n$ sets over $|S|=s$ elements, we construct a graph $G=(V,E)$ containing $|V|=s+n$ nodes - one node for each subset, and one node for each element. The graph is fully connected ($|E|=|V|\times|V|$). Let $|N|=n$ and $|M|=s$ be the sets of nodes for subsets and elements, respectively. We define the distance matrix between elements in the graph (i.e. weights on the edges) to be as follows:
$$M(i,j) = \begin{cases} \delta-1 &\mbox{if } i \in N \text{ and } j \in M \\
0 & \mbox{\text{if both }} i \in N \text{ and } j \in N \\
\delta + 1 & \mbox{if } i \in M \end{cases}$$
In other words, each node representing a subset is connected to itself and the other subset nodes with an edge of weight $0$, and to the respective node of each element it comprises with an edge of weight $\delta-1$. Element nodes are connected to all nodes with edges of weight $\delta + 1$. This construction takes polynomial time. Note that the distance of any element in $N$ (representing subsets) to itself is $0$, and the distance of every element in $M$ to itself is $\delta+1$. Let us assume we have a $c$-approximating algorithm for the representative selection problem with respect to representative set size. Any solution obtained by this algorithm with parameter $\delta$ would also yield a $c$-approximation for the set cover problem. Let us observe any result of such an algorithm - it would not return any nodes representing elements (because they are $>\delta$ distant from any node in the graph including themselves). The distance between any nodes representing subsets in the graph is $0$, so a single subset node enough is sufficient to cover all of $N$. Therefore, the representative set will only comprise elements from $N$, which directly cover elements in $M$. An optimal solution for the representative selection algorithm will also serve as an optimal solution for the original set cover problem, and vice versa (otherwise a contradiction ensues). Therefore, a $c$-approximation (with respect to set size) for the representative selection problem would also mean a $c$-approxmiation for the set cover problem. However, it is known that no approximation better than $c\mbox{log}n$ is possible \cite{raz1997sub}. Therefore, a $c$-approximating algorithm for the representative selection problem (with respect to set size) cannot be obtained unless $P=NP$.
\end{proofsketch}

\begin{mydef2}
For representative sets of optimal size $k$,\footnote{In fact, this proof applies for any value of $k$ that cannot be directly manipulated by the algorithm.} no constant-factor approximation exists with respect to the maximal distance between the optimal representative set and the samples.
\end{mydef2}

\begin{proofsketch}
We show this via a reduction from the dominating set problem. Given a graph $G=(V,E)$, a dominating set is defined as a subset $V^* \subset V$ so that every node $v \in V$ that's not in $V^*$ is adjacent to at least one member of $V^*$. Finding the minimal dominating set is known to be NP-complete \cite{gary1979computers}. 

Assume we are given a graph $G$ and are required to find a minimal dominating set. Let us generate a new graph $G'=(V,E')$, where $V$ are the original nodes of $G$ and the graph is fully connected: $|E|=|V| \times |V|$. The weights on the edges are defined as follows:

$$M(i,j) = \begin{cases} (\delta-1) &\mbox{if } (i,j) \in E \text{ (original graph) } \\
0 & \mbox{i=j} \\
2 \cdot c \cdot (\delta-1) & \mbox{otherwise} \end{cases}$$

This reduction is polynomial. Let us consider an optimal representative set with parameter $\delta$ for $G'$. Assume it is of size $k$. This would imply there's a dominating set of size $k$ which is the minimal dominating set obtaining in graph $G$. This dominating set is minimal, otherwise the representative selection set would not be optimal. Let us assume we have an algorithm for representative selection that's $c$-approximating with respect to maximal distance. If there is a dominating set of size $k$, it would imply a guarantee of $c \cdot (\delta - 1)$ on the maximal distance, implying the algorithm would behave the same as an optimal algorithm (since it cannot use edges of weight $2 \cdot c \cdot (\delta-1)$). For this reason, a $c$-maximum-distance approximation algorithm for the representative selection problem could be used to solve the dominating set problem. Since this problem is NP-hard, it implies no such approximation algorithm exists unless $P=NP$.

\end{proofsketch}

\subsection{Efficiency of $\delta$-Medoids}

The actual run time of the algorithm is largely dependent on the data and the choice of $\delta$. An important observation is that at each iteration, each sample is only compared to the current representative set, and a sample is introduced to the representative set only if it is $>\delta$ away from all other representatives. After each iteration, the representatives induce a partition to clusters and only samples within the same cluster are compared to one another. A poor choice of $\delta$, for instance $\delta < min\{d(x_i,x_j)| x_i, x_j \in S\}$ would cause all the samples to be added to the representative set, resulting in a runtime complexity of $O(|S|^2)$. In practice, however, since we only compare from samples to representatives and within clusters, for reasonable cases, we can get considerably better runtime performance. For instance, if the number of representatives is close to $\sqrt{|S|}$, the complexity would be reduced to $|S|^{1.5}$, which results in a significant speed-up. Again, note that in each iteration of the algorithm, after the partitioning phase (the \emph{RepAssign} subroutine in Algorithms $2$ and $3$) the algorithm maintains a legal representative set, so in practice we can halt the algorithm before convergence, depending on need and resources.

\section{Calculating the Distance Measures}
In this section we describe in some detail how the distance measures we used were computed, as well as some of the considerations that were involved in their formulation.

\subsection{Musical Segments Distance}

\subsubsection{Segment Information}

Every segment is transposed to C. Then, the following information is extracted from each segment:

\begin{itemize}
\item {\em Pitch Sequence} - the sequential representation of pitch frequency over time.
\item {\em Pitch Bag} - a ``bag'' containing all the pitches in the sequence, with sensitivity to registration.
\item {\em Pitch Class Bag} - a ``bag'' containing all the pitches in the sequence, {\em without} sensitivity to registration.
\item {\em Rhythm Bag} - a ``bag'' containing all rhythmic patterns in the sequence. A rhythmic pattern is defined, for simplicity, as pairs of subsequent note durations in the sequence.
\item {\em Interval Bag} - a ``bag'' containing all pitch intervals in the sequence.
\item {\em Step Bag} - a ``bag'' containing all one-step pitch differences in the sequence. this is similar to intervals, only it is sensitive to direction.
\end{itemize}

\subsubsection{Segment Distance}

We devise a fairly complex distance measure between any two musical segments, $seg1$ and $seg2$. Several factors are taken into account:

\begin{itemize}
\item {\em Global alignment} - the global alignment score between the two segments. This is calculated using the Needleman-Wunsch\cite{needleman1970general} algorithm.
\item {\em Local alignment} - the local alignment score between the two segments. This is calculated using the Smith-Waterman\cite{SmithWaterman} algorithm.
\item {\em rhythmic overlap} - the extent to which one-step rhythmic patterns in the two segments overlap.
\item {\em interval overlap} - the extent to which one-step interval patterns in the two segments overlap.
\item {\em step overlap} - the extent to which melodic steps in the two segments overlap.
\item {\em pitch overlap} - the extent to which the pitch sets in the two segments overlap. This measure is sensitive to registration.
\item {\em pitch class overlap} - the extent to which the pitch sets in the two segments overlap. This measure is invariant to registration.
\end{itemize}

The two alignment measures are combined to a single alignment score. The other measures were also combined to a separate score, which we name the {\em bag distance}. The two scores were combined using the $l2$ norm as follows: $$score_{alignment} = alignment_{global}^2 + 2 \cdot alignment_{local}^2$$
$$score_{bag} = score_{rhythmic}^2 + score_{interval}^2 + score_{step}^2$$ $$ + score_{pitch}^2 + score_{pitchClass}^2$$
$$distance = \sqrt{10 \cdot score_{bag} + score_{alignment}}$$

\subsubsection{Substitution Function}

For both the local alignment and the global alignment we used a simple exponentially attenuating function based on frequency distance to characterize the likelihood for swaps between any two notes. The function is defined as follows: $$cost(A,B) = \begin{cases} 1, & |A-B|=\mbox{3rd} \\1, & |A-B|=\mbox{5th} \\ 1.3^{\frac{|Pitch_{mid}(A)-Pitch_{mid}(B)|}{4}} , & \mbox{otherwise} \end{cases}$$
The price of introducing gaps was fixed at $1.5$.

\subsubsection{Bag Distance}
To get the bag distance score between two bags we use the calculation $\frac{|Bag_1 \Delta Bag_2|}{|Bag_1 \cup Bag_2|}$.

\subsubsection{Example}

Two example segments are given in Figure $8$ in musical notation and in Figure $9$ as midi pitch over time.

\begin{figure}[htb]
  \label{fig33}
  \centering
    \includegraphics[width=0.8\textwidth,natwidth=310,natheight=322]{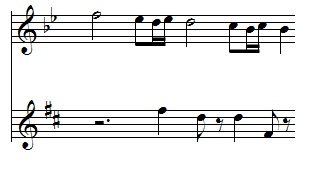}
  \caption{Two segments for example, in musical notation.}
\end{figure}

\begin{figure}[htb]
  \label{fig44}
  \centering
    \includegraphics[width=0.8\textwidth]{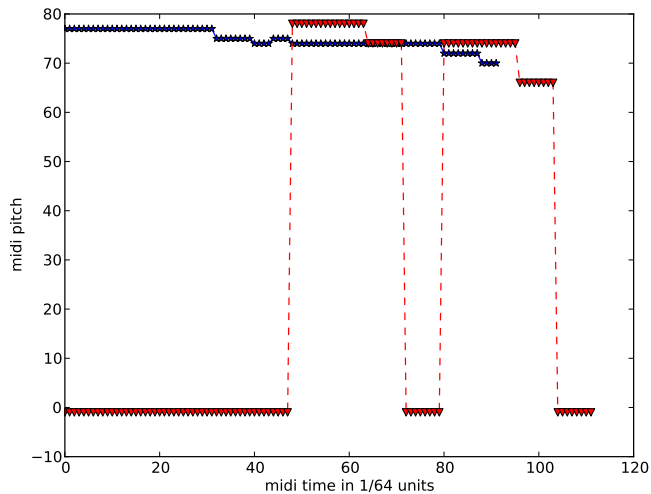}
  \caption{Same two segments, plotted as midi pitch over time.}
\end{figure}

The local alignment distance for these two segments is $0.032$.
The global alignment distance for these two segments is $1.5$.
The bag distance score for these two segments is $2.03$
The overall combined distance for these two segments after weighting is $20.4$.

\subsection{Movement Segments Distance}

\subsubsection{Segment Information}

Each segment is comprised of one agent's $(x,y)$ coordinates for $10$ consecutive timestamps. Then, each segment is translated to start from coordinates $(0,0)$, and rotated so that for all segments all players are facing the same goal. In addition to maintaining the coordinate sequence, from each such segment we extract a bag of movement-turn pairs, where the movement represents distance turns are qunatized into 6 angle bins: forward ($-30$ - $+30$ degrees), upper right ($+30$ - $+90$ degrees), lower right ($+90$ - $+150$ degrees), backwards ($-150$ - $+150$ degrees), lower left ($-90$ - $-150$ degrees), and upper left ($-30$ - $-90$ degrees). For instance, the coordinate sequence $(0,0), (0,10), (5,10), (8,14)$ induces two movement-turn elements: $10+$upper-right-turn, $5+$upper-left-turn.

\subsubsection{Segment Distance}

Given two trajectories, one can compare them as contours in 2-dimensional space. We take an alignment-based approach, with edit step costs being the RMS distance between them. Our distance measure is comprised of three elements:

\begin{itemize}
\item {\em Global Alignment} - The global alignment distance between the two trajectories once initially aligned together (that is, originating from $(0, 0)$ coordinates), calculated by the Needleman-Wunsch algorithm.
\item {\em Local Alignment} - The local alignment distance between the two trajectories, calculated by the Smith-Waterman Algorithm.
\item {\em Movement-Turn bag of words distance} - we compare the bag distance of movement-turn elements. We quantize distances into a resolution of $5$ meters to account for variation.
\item {\em Overall $\Delta$-distance and $\Delta$-angle distance} - We also consider the overall similarity of the segments in terms of total distance travelled (and the direction of the movement).
\end{itemize}

The scores are combined as follows:

$$score_{align} = alignment_{global}^2 + 2.5*alignment_{local}^2$$
$$score_{Overall \Delta} = \Delta-distance^2 + (10 \cdot \Delta-angle)^2$$
$$distance = \sqrt{100 \cdot score_{bag} + score_{align}} + score_{Overall \Delta}$$

\subsubsection{Substitution Function}
To get the substitution cost for the two alignment algorithms, we simply use the RMS distance between the two coordinates we are comparing. Given two points $P_1 = (x_1, y_1)$ and $P_2 = (x_2, y_2)$ the distance is simply $D(P_1,P_2) = \sqrt{(x_1-x_2)^2 + (y_1-y_2)^2}$. Gaps were greatly penalized with a penalty of $100$ because gaps create discontinuous (and therefore physically impossible) sequences.

\subsubsection{Bag Distance}
To get the bag distance score between two bags we use the calculation $\frac{|Bag_1 \Delta Bag_2|}{|Bag_1 \cup Bag_2|}$.

\subsubsection{Example}

Two example segments are given in Figure $10$.


\begin{figure}[htb]
  \label{fig55}
  \centering
    \includegraphics[width=0.8\textwidth,natwidth=310,natheight=322]{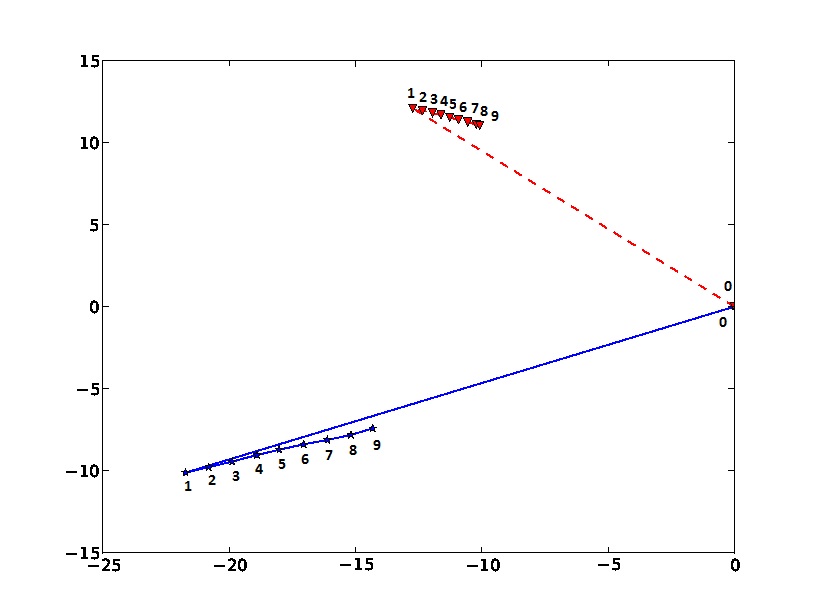}
  \caption{Two movement segments. Each coordinate in the trajectory is labelled with its timestamp in the trajectory $\in [0..9]$. Both segments begin with a long sprint towards one direction and then a sequence of small steps in the opposite direction (scales are $\times 10$).}
\end{figure}

The local alignment distance for these two segments is $20$.
The global alignment distance for these two segments is $192.7$.
The overall delta distance and angle score for these two segments is $61.66$
The overall combined distance for these two segments after weighting is $258.72$.

\end{document}